\documentclass[10pt,twocolumn,letterpaper]{article}

\usepackage{cvpr}
\usepackage{times}
\usepackage{epsfig}
\usepackage{graphicx}
\usepackage{amsmath}
\usepackage{amssymb}
\usepackage{multirow}
\usepackage{adjustbox}
\usepackage{tabularx}

\newcommand{\minisection}[1]{\vspace{0.04in} \noindent {\bf #1}\ \ }

\usepackage[pagebackref=true,breaklinks=true,letterpaper=true,colorlinks,bookmarks=false]{hyperref}

\cvprfinalcopy

\ifcvprfinal\fi
\begin{document}

\title{Learning Metrics from Teachers: Compact Networks for Image Embedding}

\author{Lu Yu$^{1,2}$, Vacit Oguz Yazici$^{2,3}$, Xialei Liu$^{2}$, Joost van de Weijer$^{2}$, Yongmei Cheng$^{1}$, Arnau Ramisa$^{3}$\\
$^{1}$ School of Automation, Northwestern Polytechnical University, Xi'an, China \\
$^{2}$ Computer Vision Center, Universitat Autonoma de Barcelona, Barcelona, Spain\\
$^{3}$ Wide-Eyes Technologies, Barcelona, Spain\\
{\tt\small \{luyu,voyazici,xialei,joost\}@cvc.uab.es, chengym@nwpu.edu.cn, aramisa@wide-eyes.it}
}

\maketitle
\begin{abstract}
   Metric learning networks are used to compute image embeddings, which are widely used in many applications such as image retrieval and face recognition. In this paper, we propose to use network distillation to efficiently compute image embeddings with small networks. Network distillation has been successfully applied to improve image classification, but has hardly been explored for metric learning. To do so, we propose two new loss functions that model the communication of a deep teacher network to a small student network. We evaluate our system in several datasets, including CUB-200-2011, Cars-196, Stanford Online Products and show that embeddings computed using small student networks perform significantly better than those computed using standard networks of similar size. Results on a very compact network (MobileNet-0.25), which can be used on mobile devices, show that the proposed method can greatly improve Recall@1 results from 27.5\% to 44.6\%. Furthermore, we investigate various aspects of distillation for embeddings, including hint and attention layers, semi-supervised learning and cross quality distillation. \footnote{Code is available at \url{https://github.com/yulu0724/EmbeddingDistillation}.}

\end{abstract}

\section{Introduction}\label{intro}

Deep neural networks obtain impressive performance for many computer vision applications, some of which have subsequently been turned into products for the general population. However, the applicability of these techniques is often limited by their high computational cost. To reduce network traffic and server costs, as well as for scalability, it is desirable to place as much of the computation as possible on the end-user side of the application. However, this is often a mobile device with limited computing power and battery life, and thus cannot compute large networks in real-time. This creates a strong demand for methods that transfer the knowledge from large networks to smaller ones, but without a significant drop in performance. 

One important class of deep networks learns feature embeddings. To be  successful, feature embeddings must preserve semantic similarity, i.e. items deemed similar by users must be close in the embedding space, despite significant visual differences such as point of view, illumination, or image quality. To bridge this gap between the semantic and visual domains, pairs or triplets of related and unrelated items are used to teach the network how to organize the output embedding space~\cite{chopra2005learning,wang2014learning,hoffer2015deep}. Embeddings were found efficient on the tasks of out-of-distribution detection~\cite{masana2018metric} and transfer learning~\cite{scott2018adapted}.
Furthermore, embedding networks are essential for computer vision, as evidenced
by the large variety of tasks in which they are used, including feature-based object retrieval~\cite{gordo2016deep}, face recognition~\cite{schroff2015facenet},  feature matching~\cite{choy2016universal}, domain adaptation~\cite{sener2016learning}, weakly supervised learning~\cite{wang2015unsupervised},  ranking~\cite{wang2014learning}, or zero-shot learning~\cite{oh2016deep}. 

Large networks are known to provide excellent feature embeddings~\cite{razavian2014cnn,schroff2015facenet}, but are often impractical for real-life applications, as mentioned before. To obtain efficient neural networks, research has focused on two main research directions: network compression and network distillation. Network compremossion reduces the number of parameters in the network~\cite{lecun1990optimal,han2015deep}, while network distillation uses a teacher-student setup in which a, typically large, teacher network is used to guide a small student network~\cite{bucilua2006model,hinton2015distilling}. This is done by using a loss function that minimizes cross-entropy between the outputs of the student and teacher network for classification. The main idea underlying network distillation is that uncertainty in the estimates of the teacher network, e.g. about whether an image contains a cat and dog, provides relevant information for the student. There are several differences between network compression and knowledge distillation. First of all, the underlying assumption of network compression is that the knowledge of the network is in the weights, whereas knowledge distillation assumes that the knowledge of the network is in the activations which arise from particular data. A second important difference is that compression algorithms typically end up with a similar network architecture than the initial large network but with less parameters (i.e. same number of layers, and layer types). In contrast, network distillation puts no restrictions on the student network design. Therefore, we focus on network distillation techniques for the efficient computation of feature embeddings with small networks. 

In this paper, we use network distillation to obtain efficient networks to learn feature embeddings. We propose two different ways of teaching metrics to students: one based on an \emph{absolute teacher}, where the student aims to produce the same embeddings as the teacher, and one based on a \emph{relative teacher}, where the teacher communicates only the distances between pairs of data points to the student. Using the CUB-200-2011 (birds), Cars-196 and Stanford Online Products datasets, we show that network distillation can significantly improve retrieval performance compared to directly training the student network on the data. We also found that the relative teacher consistently outperforms the absolute teacher. We evaluate various aspects of knowledge distillation, namely the usage of hint and attention layers, and the possibility to train from unlabelled data. We also show that a teacher with access to high-quality images can be used to improve embeddings learned with a student network with access to low-quality images. 

\section{Related work}\label{sec:related}
There is a large number of works on metric learning, see for example the survey~\cite{kulis2013metric}. Here we focus on metric learning using deep networks.

\minisection{Metric Learning} Initially metric learning with deep networks was based on Siamese architecture with contrastive loss~\cite{chopra2005learning}. Later Triplet networks were proposed which allow more local modifications of the embedding space, and do not require that all the observations of the same class collapse to the same point~\cite{hoffer2015deep,wang2014learning}. 
The progress of Siamese and Triplet networks has been hampered because of the pair (or triplet) sampling problem which arises from the huge potential space from which pairs can be sampled. For instance, in a dataset with $N$ samples, $N^{2}$ pairs could be possibly sampled, and it is therefore unfeasible to consider all of them. Therefore, hard negative mining was proposed to focus only on the pairs which induced the highest loss~\cite{simo2015discriminative}, with the expectation that the network would learn the most from them. Unfortunately, this led to severe overfitting in many cases, and semi-hard negative mining was introduced as a solution~\cite{schroff2015facenet}.
However, both hard and semi-hard negative mining have a high computational cost, which led several authors to limit the hard negative mining process to the current mini-batch~\cite{liu2017rankiqa, sohn2016improved,oh2016deep,wang2015unsupervised}. 

\minisection{Network Distillation}
 Bucila et al.~\cite{bucilua2006model} compress a large network into a small one. Their method aims to approximate a large teacher network with a single fast and compact student network. This was further improved by Hinton et al.~\cite{hinton2015distilling} by moving the teacher signal from the logits (just before the softmax) to the probabilities (after the softmax), and introducing temperature scaling to increase the influence of small probabilities. With these improvements, they achieved some surprising results on MNIST, and also showed that the acoustic model of a heavily used commercial system could be significantly improved by distilling the joint knowledge of an ensemble of models into a single one. 
FitNet~\cite{Romero15-iclr} introduced hint layers with additional losses on intermediate layers of the network to communicate knowledge of the teacher to the student. They show that this helps to train deep and thin networks, which cannot be trained from scratch without teacher supervision. In addition, these students can outperform the teacher network while using less memory. Zhang et al.~\cite{zhang2017deep} show that a group of students, without a teacher, which are jointly trained with similar losses as between teacher and student, can outperform standard ensemble learning. Network distillation can also be used to compress multiple teachers into a single student network~\cite{gao2017knowledge}. Most literature on network distillation focuses on image classification, but recently several works have investigated applying the theory to object detection~\cite{chen2017learning} and pedestrian detection~\cite{shen2016teacher}. 

Only two works have previously addressed knowledge distillation for embeddings. Chen et al.~\cite{chen2017darkrank}  bring the 'learning to rank' technique into deep metric learning for knowledge transfer. It is formalized as a rank matching problem between teacher and student networks. Their list-wise loss can easily overflow due to the product operation when computing the probability of the permutation, which severely limits the batch size that can be used. PKT~\cite{passalis2018learning} applies a different approach where they model the interactions between the data samples in the feature space as a probability distribution. In our experiments we show that our proposed relative teacher outperforms the DarkRank and PKT significantly.

\section{Preliminaries}\label{sec:prel}
In this paper, we will apply network distillation to metric learning networks. This section will briefly introduce both. 

\begin{figure}[tb]
\begin{center}
\includegraphics[width=0.4\textwidth]{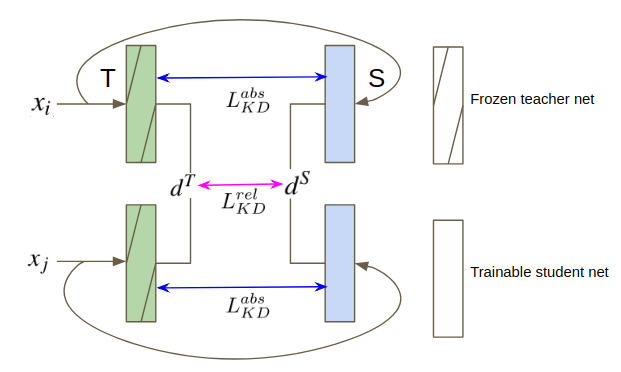}
\caption{Graphical illustration of the two knowledge distillation losses we propose for metric learning. $L_{KD}^{abs}$ aims to minimize the distance between the student and teacher embedding of the same image. $L_{KD}^{rel}$ compares the distance in the embedding of the teacher between two images, with the distance of the same two images in the student embedding. It aims to make the two distances as similar as possible.} 
\label{fig:framework}
\end{center}
\end{figure}

\subsection{Metric Learning}
A fundamental step in most computer vision applications is transforming the initial representation of the images (i.e.~pixels) into another one with more desirable properties. This process is often denoted as \textit{feature extraction}, and projects the images to a high-level representation that captures the semantic characteristics relevant to the task. How images are organized in this high-level representation is crucial to the success of many applications. For example, image retrieval, k-NN or Nearest Class Mean classifiers are directly based on the distances between these high-level image representations. Metric learning addresses this problem and intends to map the input feature representations to an embedding space where the L2 distance correlates with the desired notion of similarity. In this work we will focus on deep, or end-to-end, metric learning, where the whole feature extraction network is trained jointly to generate the best possible representation.

Siamese networks map data to an output space where distance represents the semantic dissimilarity between the images~\cite{bromley1994signature,chopra2005learning}. Triplet networks were 
proposed by Hoffer et al.~\cite{hoffer2015deep} based on the work of Wang et al.~\cite{wang2014learning}. In contrast to Siamese networks, they use triplets formed by an anchor ($x_{a}$), a positive instance ($x_{p}$) and a negative instance ($x_{n}$), as input. The anchor and the positive instances correspond to the same category, while the negative instance is from a different one. The objective is to guarantee that the negative instance is further away from the anchor than the positive (plus a margin $m$). The Triplet loss is given by:
\begin{equation}
L_{T}= \max(0, d_+-d_- + m),\label{eq:triple}
\end{equation}
where $d_+$ and $d_-$ are the distances between the anchor and the positive and negative instances respectively. The Triplet network imposes only \emph{local} constraints on the output embedding, which can simplify convergence when compared to the Siamese network, reportedly harder to train. In the experiments we will show results of network distillation for  embeddings learned with triplet losses.

\subsection{Network Distillation}

\begin{figure*}[tb]
\begin{center}

  \includegraphics[width=0.8\textwidth]{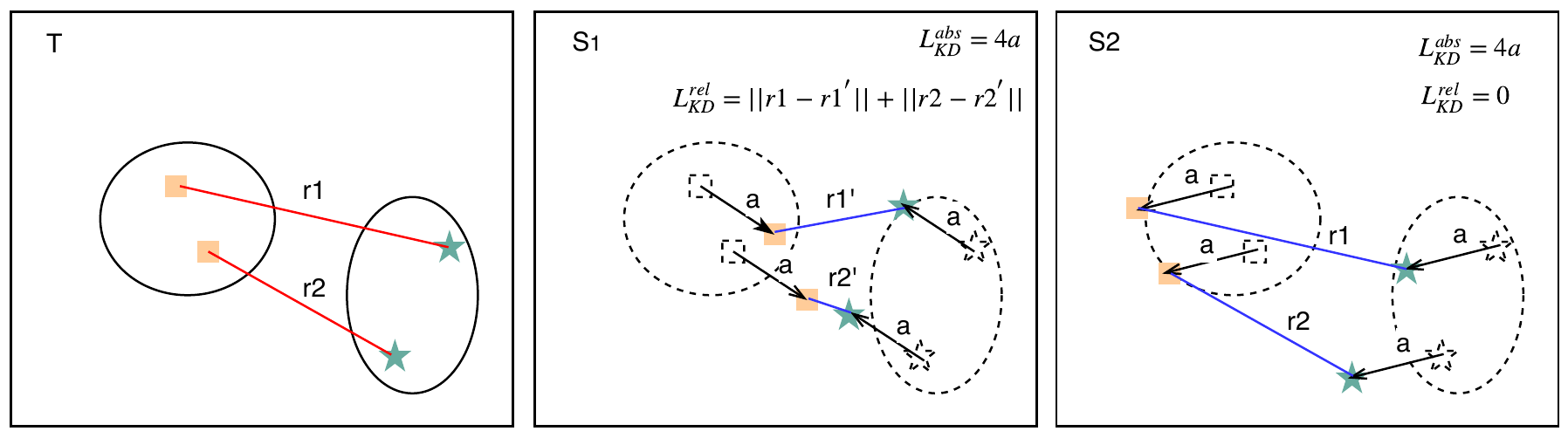}
  \caption{Illustration of difference between absolute and relative teacher. (left) Example of four data points in the embedding space of teacher. We consider two samples from two classes (indicated by square and star). (middle and right) show the absolute and relative loss for two student embeddings $S1$ and $S2$ (the teacher location of the points is given in dashed lines). The (right) embedding is preferable since it is exactly equal to the teacher (except for a translation). This is only appreciated by the relative teacher, whereas the absolute teacher assigns equal loss to both.}
  \label{fig:space}
\end{center}
\end{figure*}

Network distillation~\cite{hinton2015distilling,Romero15-iclr} aims to transfer the knowledge of a large teacher network $T$ to a small student network $S$. The objective for network distillation for classification networks is defined as:
\begin{equation}
L_{KD}=H(y_{true},P_{\tau=1}^{S})+\lambda H(P_{\tau }^{T},P_{\tau }^{S}),
\end{equation}
where $\lambda$ is used to balance the importance of two cross-entropy losses $H$: the first one corresponds to the traditional loss between the predictions of the student network and the ground-truth labels $y_{true}$, and the second one between the \emph{annealed} probability outputs of the student and teacher networks. This loss encourages the student to make similar predictions as the teacher network. The information of the teacher $P_{\tau }^{T}$ could be more valuable than the ground truth $y_{true}$ for the student network, because it also contains information of which classes could possibly be confused with the true label for a particular image. More precisely, $P_{\tau }^{T}$ and $P_{\tau }^{S}$ are:
\begin{equation}
P_{\tau }^{T}={\rm{softmax}}(\frac{a_{T}}{\tau}), \; P_{\tau }^{S}={\rm{softmax}}(\frac{a_{S}}{\tau}),
\end{equation}
where $a_S$ and $a_T$ are the (pre-softmax) activations of the student and teacher networks respectively, and temperature $\tau$ is a relaxation which is introduced to soften the signal arising from the output of the networks. It was found that for complex classification tasks $\tau=1$ obtained good results~\cite{chen2017learning}. $P_{\tau=1}^{S}$ is equal to the  output of the standard student network without any temperature scaling. 

\section{Distillation for Metric Learning}\label{sec:dist}
Wide and deep networks with large amounts of parameters are known to obtain excellent results~\cite{Simonyan2015}, however they are very time consuming and memory demanding. Network distillation is proven to be one of the solutions to handle this problem in the classification field~\cite{hinton2015distilling}. 
In this section we extend the theory of knowledge distillation to networks that aims to project images into an embedding space. In addition we will discuss the incorporation of hint and attention transfer between student and teacher. 

\subsection{Knowledge distillation for embedding networks}\label{KD}
Traditional network distillation has focused on networks which perform classification, and are trained with a cross-entropy loss~\cite{hinton2015distilling,Romero15-iclr}. During training the output class distribution produced by the student is enforced to be close to that of the teacher. This is shown to obtain much better results than directly training the student on the available data; the main reason for this performance difference is that confusions between classes of the teacher reveal relevant information to the student, thereby providing a richer training signal than ground truth labels would provide~\cite{hinton2015distilling}. 

Here we extend the technique of knowledge distillation to networks that are used to project input data into an embedding (from now on called \emph{embedding networks}). These embeddings are then typically used to perform distance computation. For example, to provide a ranked list of similar data (ordered according to the distance). For knowledge distillation it is important to consider what is the knowledge that is contained in the embedding network. One could consider the actual embedding (meaning the coordinates of the embedding) to be the knowledge of the network. Another point of view would be to consider the distances which are computed based on the embedding network to be the actual knowledge, since this is actually the main purpose of the embedding network. We will consider both these points of view and design two different teachers: one teacher, called \emph{absolute teacher} which teaches the exact coordinates to the student and one teacher, called \emph{relative teacher} which only teaches the distance between data pairs to the student. 

In the first approach, the absolute teacher, we directly minimize the distance between the student ($F^S$) and teacher ($F^T$) embeddings. This is done by minimizing:
\begin{equation}\label{eq:abs_loss}
L_{KD}^{abs}=\left\| {F^S \left( {x_i } \right) - F^T \left( {x_i } \right)} \right\|,
\end{equation}
where $\left\| . \right\|$ refers to the Frobenius norm.

As a second approach, we consider the relative teacher, which enforces the student network to learn any embedding as long as it results in similar distances between the data points. This is done by minimizing the following loss:
\begin{equation}
L_{KD}^{rel}=  \left | d^S -  d^T \right |,
\label{eq:dist}
\end{equation}
where $d^S$ and $d^T$ are, respectively, the distances between the student and teacher embeddings of images $x_i$ and $x_j$:
\begin{equation}\label{kd1}
\begin{array}{l}
 d^S  = \left\| {F^S \left( {x_i } \right) - F^S \left( {x_j } \right)} \right\|, \\ 
 d^T  = \left\| {F^T \left( {x_i } \right) - F^T \left( {x_j } \right)} \right\|, 
 \end{array}
\end{equation}
The minimization loss in Eq.~\ref{eq:dist} is equal to the loss used in the classical problem of multidimensional scaling (MDS)~\cite{cox2000multidimensional}. There the dissimilarities between points is known and the goal is to find coordinates for the points in some (low-dimensional) space where the dissimilarities between the points is equal to their dissimilarity.

\begin{figure*}[tb]
	\begin{center}
		\includegraphics[width=0.8\textwidth]{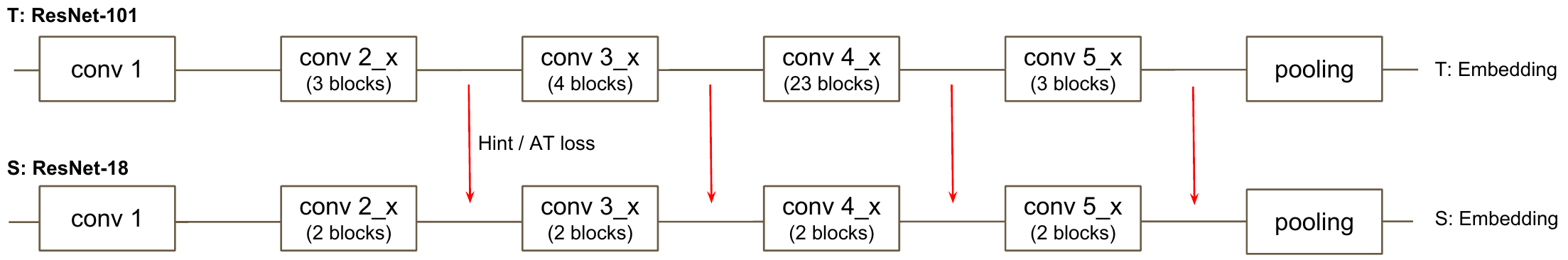}
		\caption{Schematics of teacher-student hint/attention transfer.}
		\label{fig:hint_AT}
	\end{center}
\end{figure*}
	
A graphical illustration which shows the relevant distances that are used by the absolute and relative teacher is provided in Fig.~\ref{fig:framework}. The teacher networks are frozen during training of the student network. The absolute teacher minimizes the distance between the student and teacher embedding for each training sample. In case of the relative teacher, one should consider pairs of data points, since during training the student network is optimized 
to obtain similar distances between instances of data points. 

As reported by several other authors ~\cite{hinton2015distilling,zhang2017deep}, we train the student network by simultaneously considering the standard metric learning loss $L_{ML}$ (see Eq.~\ref{eq:triple}) and the loss $L_{KD}^{T}$ imposed by the teacher, according to: 
\begin{equation}
L = L_{ML} + \lambda L_{KD}^{T}, 
\end{equation}
where $T \in \left\{ {abs,rel} \right\}$ and $\lambda$ is a trade-off parameter between the different losses, which is learned by cross validation. 

In Fig.~\ref{fig:space} an illustration of the two distillation losses is given for two different student embeddings, indicated by $S1$ and $S2$. The embedding $S2$ is preferable because it is equal to the teacher embedding except for a translation which does not influence the ranking of data points. The S1 embedding actually changes the relation between samples, and would not obtain similar results as the teacher network. However, if we consider the absolute loss for these two scenarios we see it assigns equal loss to both embeddings. The relative loss does correctly assign a lower (zero) loss to the S2 embedding. By focusing on the relevant parameter (namely the distance), we expected that relative teachers are able to better guide the student to a similar embedding than the student network. 

\subsection{Learning from hints and attention}\label{sec:hints}	

In this section we consider two techniques that have shown to improve results for distillation of classification networks. The techniques we consider are: the introduction of hint layers~\cite{Romero15-iclr} and the usage of attention~\cite{zagoruyko2016paying}. Both were proposed to improve the learning of student networks. We are interested to know if these techniques also generalize to knowledge distillation for embedding networks. 

Romero et al.~\cite{Romero15-iclr} propose to improve knowledge distillation by introducing an additional loss on intermediate representations learned by the teacher (called hints). The loss which incorporates the hint layers is given by:
\begin{equation}
L_{hint}=\left\| {F^S_{hint} \left( {x_i } \right) - F^T_{hint} \left( {x_i } \right)} \right\|,
\end{equation}
where $F^T_{hint}\in {R}^{w\times h\times d}$ where $w$, $h$ and $d$ are dimensions of the activation map of the hint layer. 

In this work~\cite{Romero15-iclr}, they first train the network until the hint loss, and then train the whole network only based on the distillation loss. In contrast, we propose to learn with both losses simultaneously, as was also done in~\cite{chen2017learning,shen2016teacher}. Combining the knowledge distillation loss of either the absolute or relative teacher we would obtain as a final objective function:
\begin{equation}\label{equ:hint}
L = L_{ML} + \lambda L_{KD}^{T} + \mu L_{hint}, 
\end{equation}
where $T \in \left\{ {abs,rel} \right\}$ and $\mu$ is used to balance the relative weight of the hint loss.  

Zagoruyko and Komodakis~\cite{zagoruyko2016paying} improve the performance of student networks by forcing them to mimic intermediate attention maps of a powerful teacher network. Attention maps convey what spatial locations in the image are considered relevant to the teacher network for its interpretation. Communicating this information can therefore guide the student network in learning the task at hand. They propose to compute activation-based spatial attention according to:
\begin{equation}
A_{sum}^{T}(x_i)=\sum_{l=1}^{C^k}\left | F^T_{kl}(x_i) \right |^2,
\end{equation}
where $F^T_{kl}(x_i)\in {R}^{w\times h}$ refers to the $l$-th map of the activation of the $k$-th layer for image $i$. Here $C^k$ denotes the number of feature maps in the $k$-th layer of the teacher net. We use $\left | . \right |$ to refer to the pixel-wise absolute value, as a results $A_{sum}^{T}(x_i)\in {R}^{w\times h}$. A similar equation is used to compute $A_{sum}^{S}(x_i)$ from the student activation maps $F^S$. The attention loss is then defined as: 
\begin{equation}
L_{AT}=\left \| \frac{A_{sum}^{T}(x_i)}{\left \| A_{sum}^{T}(x_i) \right \|_2}-\frac{A_{sum}^{S}(x_i)}{\left \| A_{sum}^{S}(x_i) \right \|_2}\right \|.
\end{equation}
This enforces the student to assign its attention to the same locations which were deemed important by the teacher.

The full objective function for the attention based metric learning network becomes:
\begin{equation}\label{equ:AT}
L = L_{ML} + \lambda L_{KD}^{T} + \kappa L_{AT},
\end{equation}
where $\kappa$ defines the relative weight of the attention loss. 

In Fig.~\ref{fig:hint_AT} we show how the hint and attention layers are incorporated between a ResNet-101 teacher and a ResNet-18 student network. Both hint and attention losses are applied on multiple layers\footnote{For the student we take the output of each block, and compare it to the last but one layer for each block of the teacher. The dimensionality of these layers is the same.}. Results for this scheme will be presented in the experimental section. 

\begin{table*}[tb]
\centering
\caption{Retrieval Performance on the CUB-200-2011 and Cars-196 dataset. 'ML':metric learning loss, 'hint':hint loss; 'AT': attention loss; KD (abs): absolute teacher loss; KD (rel):relative teacher loss }
\label{table:main}
\resizebox{1.45\columnwidth}{!}{%
\begin{tabular}{c|cccll|cccll}
\hline
                         & \multicolumn{5}{c|}{CUB-200-2011} & \multicolumn{5}{c}{Cars-196}    \\ \hline
Recall@K                 & 1     & 2    & 4    & 8    & 16   & 1    & 2    & 4    & 8    & 16   \\ \hline
Student (ResNet-18) & 51.7  & 63.7 & 74.2 & 83.7 & 90.9 & 46.7 & 59.5 & 71.6 &	82.3 & 90.6\\ \hline
PKT~\cite{passalis2018learning} 
                         & 53.1  & 64.2 & 75.4 & 84.6 & 91.6 & 46.9 &	59.9 &	72.1 & 82.8 & 90.8\\ \hline
DarkRank~\cite{chen2017darkrank} 
                         & 56.2  & 67.8 & 77.2 & 85.0 & 91.5 & 74.3 &	83.6 &	90.0 & 94.2 & 96.9 \\ \hline
ML+KD (abs)              & 54.9  & 66.5 & 76.5 & 85.0 & 91.3 & 70.6 & 80.7 & 88.0 & 93.2 & 96.0\\
ML+KD (rel)              & \textbf{58.0}  & \textbf{69.0} & \textbf{79.4} & \textbf{87.8} & \textbf{93.6} & \textbf{76.6} & \textbf{85.4}	& \textbf{91.2} & \textbf{95.0} &	\textbf{97.3} \\ \hline
ML+KD (abs)+hint          & 55.0  & 66.5 & 76.6 & 84.9 & 91.1 & 71.3 & 81.2 &	88.1 & 92.7	& 95.9\\
ML+KD (rel)+hint          & 57.4  & 68.8 & 79.1 & 87.4 & 93.1 & 76.4 & 85.5 &	91.3 & 95.1 & 97.2 \\ 
ML+KD (abs)+AT            & 55.0  & 66.3 & 76.9 & 85.3 & 91.8 & 71.1 & 81.3 &	88.3 & 93.1	& 96.0\\
ML+KD (rel)+AT           & \textbf{58.1}  & \textbf{69.2} & \textbf{79.6} & \textbf{85.3} & \textbf{91.3} & \textbf{76.4} & \textbf{85.7} & \textbf{91.7} & \textbf{95.0} & \textbf{97.2} \\ \hline
Teacher (ResNet-101)     & 58.9  & 70.4 & 80.7 & 88.2 & 93.5 & 74.8 & 83.6 & 89.9 & 93.8 & 96.5 \\ \hline
\end{tabular}}
\end{table*}

\section{Experimental Results}\label{sec:exp}

We show results on several benchmark datasets. Our method is implemented with the PyTorch framework~\cite{paszke2017automatic}. We will release a GitHub page with code upon acceptation. 

\subsection{Retrieval on Fine-grained Datasets}

\begin{table}[tb]
\centering
\caption{Comparison on Stanford Online Products dataset. }
\label{table:product}
\begin{tabular}{c|cccc}
\hline
                         & \multicolumn{4}{c}{Stanford Online Products} \\ \hline
Recall@K                 & 1         & 10        & 100       & 1000      \\ \hline
Student (ResNet-18) & 61.7      & 78.6      & 90.2      & 96.8      \\ \hline
ML+KD (abs)              & 68.0	 & 82.7  & 92.1 & 97.4          \\
ML+KD (rel)              & 67.7	   & 83.0      & 92.0      & 97.2         \\ \hline
Teacher (ResNet-101)     & 69.5      & 84.4      & 93.1      & 97.9      \\ \hline
\end{tabular}
\end{table}

\minisection{Datasets:}We evaluate our framework for the task of image retrieval on three fine-grained datasets:
\begin{itemize}
\item CUB-200-2011: this dataset was introduced in~\cite{wah2011caltech}. It has 200 classes with $11,788$ images in total. 
\item Cars-196: this dataset contains $16,185$ images of $196$
cars classes and was introduced in ~\cite{krause20133d}.
\item Stanford Online Products: this dataset introduced in~\cite{oh2016deep} contains $120,053$ images of $22,634$ products collected from eBay.com.
\end{itemize}

\begin{figure*}[tb]
\begin{center}
  \includegraphics[width=0.8\textwidth]{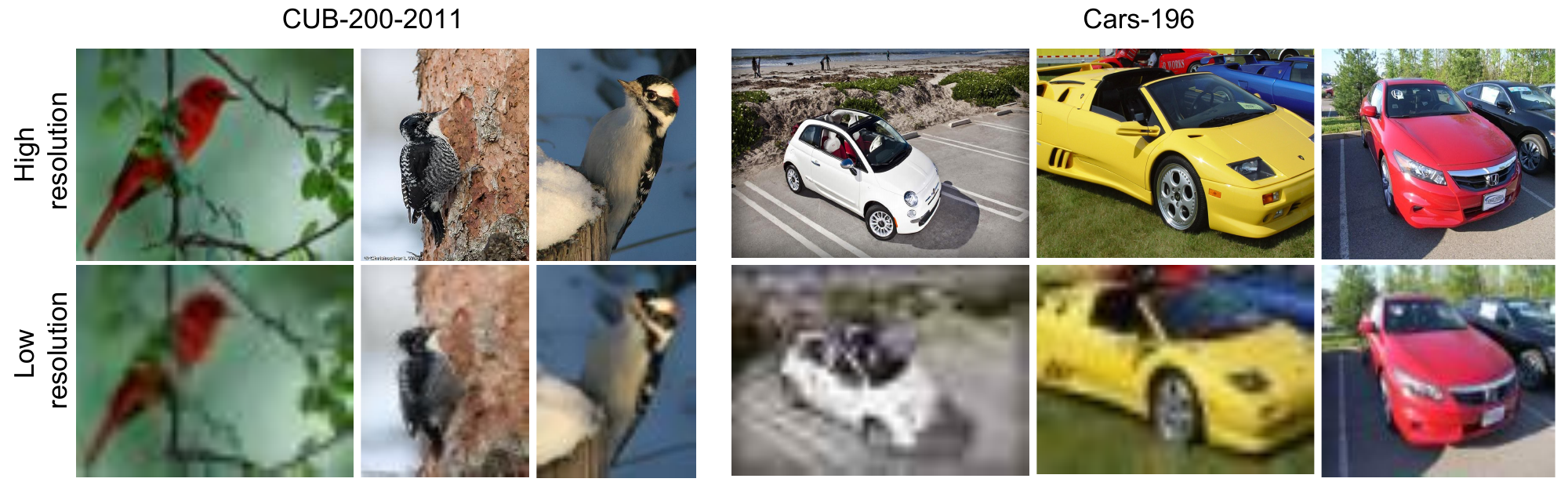}
  \caption{ Example images from two fine-grained datasets CUB-200-2011 and Cars-196 used in our experiments. The top row shows examples of high-quality images and the bottom row shows examples of the corresponding low-quality images.}
  \label{fig:dataset}
\end{center}
\end{figure*}

Example images of CUB-200-2011 and Cars-196 are shown in  Fig.~\ref{fig:dataset}. We follow the evaluation protocol proposed in~\cite{oh2016deep}. By excluding some classes from the training of the embedding, we can evaluate at testing time how good the embedding generalizes to unseen classes. Therefore, the first half of classes are used for training and the remaining half for testing. For instance, on CUB-200-2011 dataset, 100 classes (5,864 images) are for training and the remaining 100 classes (5,924 images) are for testing. We divide the training set into 80\% as training and 20\% as validation. 

\minisection{Experimental Details:}For these experiments, we use a ResNet-101 as the teacher network and a ResNet-18 as the student network (see also Fig.~\ref{fig:hint_AT}). The comparison of the number of parameters of these two networks is shown in Table~\ref{tab:comp_mobile}. After the average pooling layer, a linear 512-dimensional embedding layer is added and the triplet loss is used for training both teacher and student networks. The Adam~\cite{kingma2015j} optimizer is used with a learning rate of $1e^{-5}$, and a mini-batch of 32 images. We apply hard negative mining~\cite{simo2015discriminative} on triplet loss. For preprocessing, we follow the previous work~\cite{opitz2017bier}, we resize all images to 256$\times$256, and crop 224$\times$224 patches randomly. Horizontal flip is used for data augmentation. We fine-tune both student and teacher networks from pre-trained ImageNet models with the same preprocessing. During test time, we only use the 224$\times$224 pixel center crop to predict the final feature representation used for retrieval. The optimal parameters are selected according to the performance on the validation set for all of the experiments. We use the whole training set to retrain with optimal parameters for a fixed number of epochs.

For evaluation, we use the Recall@$K$ metric~\cite{oh2016deep}: each image in the test set is projected using the trained network, and if one of the $K$ closest images in the embedding space has the same label, it is considered as a positive result. The final score is the fraction of positive results obtained on all the test images. Furthermore, the reported results in all tables are the average over three repeated experiments.

\minisection{Baselines:} We start by considering the results of the student and teacher network in Table~\ref{table:main} and Table~\ref{table:product}.
Not surprisingly, the teacher network is able to leverage the additional capacity to learn better embeddings. On the CUB-200-2011 dataset, we obtain a $R@1$ accuracy of 58.9\% for the teacher and 51.7\% for the student. This is consistent for the other evaluated recall levels, although the gap narrows with higher $K$. This shrinking performance gap is mirrored in the Car-196 dataset, with the teacher net attaining a 28.1\% better $R@1$, and a 5.9\% better $R@16$. On the Stanford Online Products dataset, the gap between the teacher and student network is 7.8\%.

We also compare our method with the DarkRank method~\cite{chen2017darkrank} and PKT~\cite{passalis2018learning}\footnote{For these results we used the code made available by the authors.}. The experiments show that the relative teacher network significantly outperforms DarkRank and PKT
, obtaining 1.8\% and 4.9\% more on CUB-200-2011 and 2.3\% and 29.7\% on Cars-196.

\minisection{Absolute and Relative Loss:} Next we incorporate the additional knowledge distillation losses to the student metric learning objective (indicated by \emph{ML+KD}). Table~\ref{table:main} 
shows that results improve for every dataset and recall level, regardless of the loss used.  The performance improvement at Recall@1 is 3.2\% and 6.3\% respectively for the absolute and relative teacher on CUB-200-2011.  On Cars-196 we see a similar behavior, again the relative teacher is outperforming the absolute teacher. The student trained with the relative teacher has an stunning performance gain of almost 30.0\%. It is interesting to note that the relative teacher even outperforms the teacher by 1.8\% while having less parameters. On Stanford Online Product dataset, both absolute and relative teacher obtain similar results, and outperform the direct training of the student network with 6.0\%.
In conclusion, the proposed distillation methods consistently manage to improve the performance of student network, especially those trained with the relative teacher.

\begin{figure}[tb]
\begin{center}
  \includegraphics[width=0.35\textwidth]{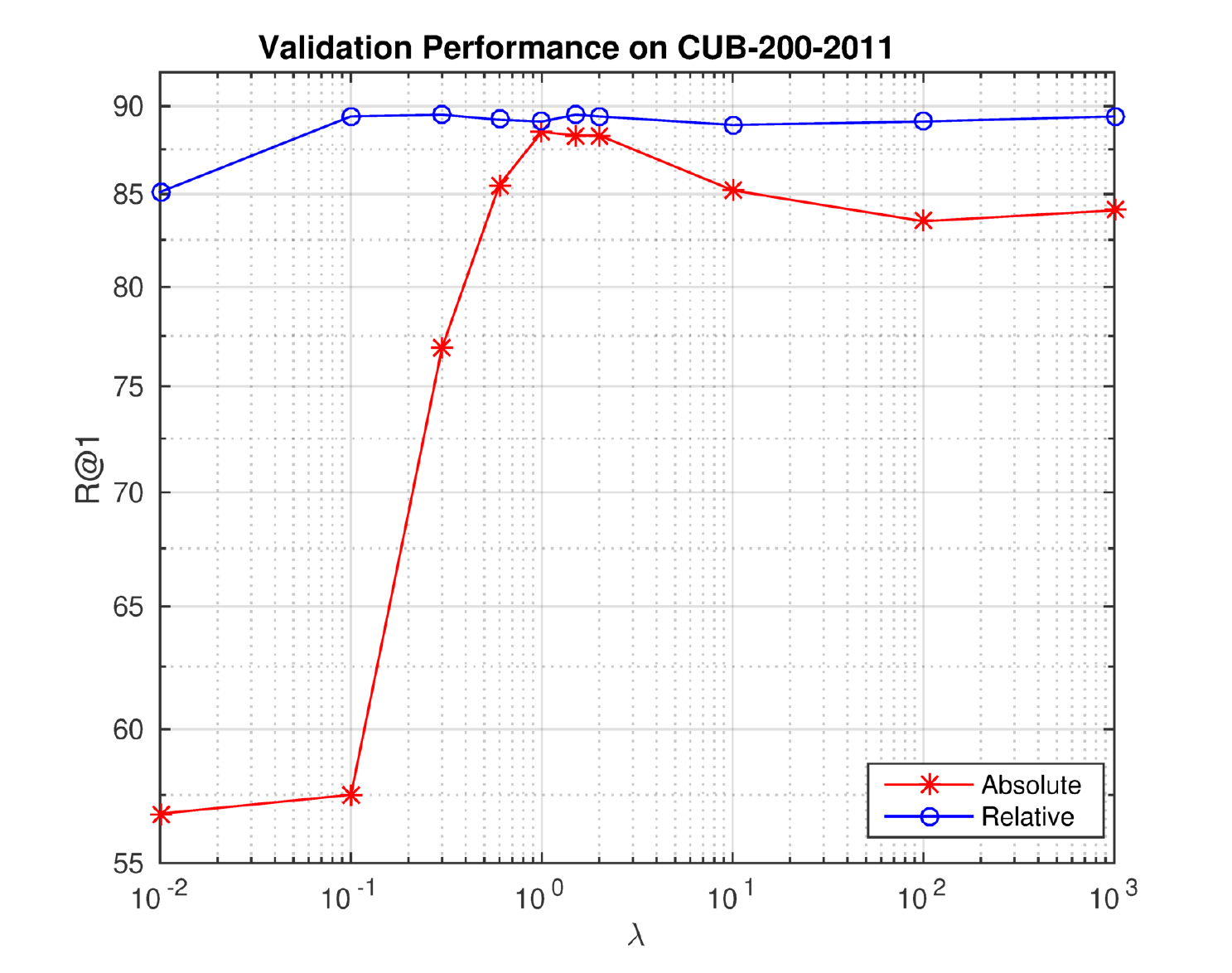}
  \caption{R@1 as a function of $\lambda$ on CUB-200-2011 dataset.} 
  \label{fig:cub_lambda}
\end{center}
\end{figure}

\minisection{Hint and attention losses:} Here we investigate if hint~\cite{Romero15-iclr} and attention~\cite{zagoruyko2016paying} layers are beneficial for knowledge distillation of embedding networks (see also Section~\ref{sec:hints}). We combine them with our proposed absolute and relative losses according to Eq.~\ref{equ:hint} and Eq.~\ref{equ:AT}. The results are summarized in Table~\ref{table:main}. We found that adding a hint layer was not stable. This is probably because the hint layer is similar as the absolute teacher forcing the network to learn the exact same embedding as the teacher, and therefore only helps when combined with the absolute teacher. Adding attention layers in general provided a small gain but the gain was not as large as reported for classification networks~\cite{zagoruyko2016paying}. 

\minisection{Sensitivity to $\lambda$} As shown in Figure~\ref{fig:cub_lambda}, we compare R@1 performance as a function of different $\lambda$ values on the validation set of CUB-200-2011 for both absolute and relative teachers. It is noteworthy that the relative teacher has a stable performance on a large range of the trade-off parameter $\lambda$, while the absolute teacher only works in a very narrow range. It suggests that in practice the selection of the $\lambda$ parameter is not that essential for the relative teacher.

\begin{table}[tb]
\centering
\caption{Semi-supervised results on CUB-200-2011.}
\label{table:semi}
\resizebox{0.9\columnwidth}{!}{%
\begin{tabular}{ccccc}
\hline
\multicolumn{1}{c|}{}                                             & \multicolumn{1}{c|}{Recall@$K$}               & 1    & 2    & 4          \\ \hline
\multicolumn{1}{c|}{\multirow{3}{*}{50\% labeled}}                & 
\multicolumn{1}{c|}{Student (ResNet-18)} & 51.0 & 63.0   & 74.0   \\
\multicolumn{1}{c|}{}                                             & \multicolumn{1}{c|}{ML+KD (abs)}            & 51.7 & 63.2   & 73.9    \\
\multicolumn{1}{c|}{}                                             & \multicolumn{1}{c|}{ML+KD (rel)}            & \textbf{56.0} & \textbf{66.7}   & \textbf{77.6}   \\ 
\multicolumn{1}{c|}{}                                             &
\multicolumn{1}{c|}{Teacher (ResNet-101)}     & 58.1 & 70.0   & 80.1   \\ \hline
\multicolumn{1}{c|}{50\% labeled  } & \multicolumn{1}{c|}{(ML+KD (abs))/KD (abs)}            & 53.9 & 65.2 & 75.8  \\
\multicolumn{1}{c|}{+50\% unlabeled}                                             & \multicolumn{1}{c|}{(ML+KD (rel))/KD (rel)}            & \textbf{57.2} & \textbf{68.0} & \textbf{78.2}   \\ \hline
\multicolumn{1}{c|}{\multirow{2}{*}{50\% unlabeled}}              & \multicolumn{1}{c|}{only KD (abs)}            & 49.8 & 60.8   & 71.0  \\
\multicolumn{1}{c|}{}                                             & \multicolumn{1}{c|}{only KD (rel)}            & \textbf{55.5} & \textbf{67.0}   & \textbf{77.4}   \\ \hline
\end{tabular}}
\end{table}

\subsection{Semi-Supervised Learning}\label{sec:semi}
One of the interesting properties of network distillation is that it allows for the usage of unlabeled data. This was observed by~\cite{zhang2017deep} and we apply this idea here to distillation for embedding networks. 
The knowledge distillation losses of Eqs.~\ref{eq:abs_loss} and~\ref{eq:dist} do not require any labels. Knowing the estimation of the teacher network for unlabeled data can help the student network to better approximate the teacher network. In addition, the existing problems of pair sampling can be avoided in semi-supervised learning for the student network because for the unlabeled images we do not apply the triplet loss as it requires labels. In the experiments we evaluate the benefit of adding unlabeled data to the student network for embedding learning on the CUB-200-2011 dataset.

We randomly select half of the training images per class as labeled data, and consider the rest as unlabeled data. Thus, here we have two teacher-student learning mechanisms, one is used on the labeled training set with both the ground truth annotations and information transferred from the teacher, and the other one is applied to the unlabeled training set with only information from the teacher by means of a distillation loss. The results of this experiment can be seen in Table~\ref{table:semi}. The first row (50\% labeled) shows our approach using only the remaining labeled data, with similar performance as in the previous experiments: the performance obtained by the relative teacher is closer to teacher network and better than the absolute teacher. In the second row we add the remaining 50\% of unlabeled data. This leads to improved Recall@K with both losses, but especially for the relative teacher.

Finally, in the third row, we consider the case where we have access to a trained teacher network, but no labelled data at all, to train the student network. Here we would like to highlight the results of the relative teacher, 
since it manages to increase performance by 4.5\% compared to the student network trained with 50\% of labeled training data.

\subsection{Very Small Student Networks}

\begin{table}[tb]
\centering
\caption{Parameter Comparison of Different Networks .}
\label{tab:comp_mobile}
\resizebox{0.8\columnwidth}{!}{%
\begin{tabular}{c|ccc}
\hline
Network    & ResNet-101                   & ResNet-18                    & MobileNet-0.25              \\ \hline
Parameters & $\sim$48.1 M & $\sim$11.3 M & $\sim$0.3 M \\ \hline
\end{tabular}}
\end{table}

MobileNets~\cite{howard2017mobilenets} are efficient but light-weight networks that can be easily matched to the design requirements for mobile and embedded vision applications. To show the potential of our method on very small networks, we propose to use the  MobileNet-0.25 (0.25 is the width multiplier) network as our student network and use ResNet-101 as the teacher network. 
The number of parameters per network is given in
Table~\ref{tab:comp_mobile}. We can see that the number of parameters of MobileNet-0.25  is almost 40 times smaller than that of the previous student network (ResNet-18) and 160 times smaller than that of the teacher network (ResNet-101). 

Table~\ref{tab:025} shows retrieval performance results on CUB-200-2011 with MobileNet-0.25 as the student network. We can observe that the Recall@K for $K = 1, 2$ of the teacher network is almost 2 times higher than the student network. After our relative teacher is applied, the performance gain is 17.1\% at Recall@1 and 15.5\% at Recall@16 higher compared to the original student model. 
             
\begin{table}[tb]
\centering
\caption{Performance on CUB-200-2011 with MobileNet-0.25.}
\label{tab:025}
\begin{adjustbox}{width=0.45\textwidth}
\begin{tabular}{cccccc}
\hline
\multicolumn{1}{c|}{Recall@K} & 1 & 2 & 4 & 8 & 16 \\ \hline
\multicolumn{1}{c|}{Student (MobileNet-0.25 )} & 27.5  & 35.8 & 46.0 & 58.5  &70.6  \\ \hline
\multicolumn{1}{c|}{ML+KD (rel)} & 44.6 & 56.0 & 66.4 & 77.3 & 86.1  \\ \hline
\multicolumn{1}{c|}{Teacher (ResNet-101)} & 58.9  & 70.4 & 80.7 & 88.2 & 93.5 \\ \hline
\end{tabular}
\end{adjustbox}
\end{table}

\begin{table}[tb]
\centering
\caption{Cross quality results on the CUB-200-2011 and Cars-196 datasets with low resolution and unlocalized object degradations.}
\label{tab:cqd}
\tabcolsep=0.08cm
\resizebox{0.99\columnwidth}{!}{%
\begin{tabular}{cc|ccc|ccc}
\hline
                                                   &                      & \multicolumn{3}{c|}{Low Resolution}                                           & \multicolumn{3}{c}{Unlocalized}                                              \\ \hline
\multicolumn{1}{c|}{}                              & Recall@$K$           & 1             & 2             & 4                        & 1             & 2             & 4               \\ \hline
\multicolumn{1}{c|}{\multirow{4}{*}{CUB-200-2011}} & Student  (ResNet-18) & 44.4          & 54.7          & 65.3                    & 43.6          & 54.5          & 66.9                    \\
\multicolumn{1}{c|}{}                              & ML+KD (abs)          & 45.7          & 56.8          & 68.3                   & 43.0          & 54.5          & 66.1                   \\
\multicolumn{1}{c|}{}                              & ML+KD (rel)          & \textbf{46.2} & \textbf{57.4} & \textbf{68.6}  & \textbf{45.9} & \textbf{57.9} & \textbf{69.3}  \\
\multicolumn{1}{c|}{}                              & Teacher (ResNet-18)  & 53.7          & 65.2          & 74.7                   & 54.8          & 67.2          & 78.7                  \\ \hline
\multicolumn{1}{c|}{\multirow{4}{*}{Cars-196}}     & Student (ResNet-18)  & 37.5          & 50.0          & 62.6            & 54.0          & 67.3          & 78.2                    \\
\multicolumn{1}{c|}{}                              & ML+KD (abs)          & 58.6          & 70.7          & 80.7                  & 57.7          & 70.4          & 80.6              \\
\multicolumn{1}{c|}{}                              & ML+KD (rel)          & \textbf{58.9} & \textbf{71.0} & \textbf{81.1}  & \textbf{61.9} & \textbf{74.4} & \textbf{84.2}  \\
\multicolumn{1}{c|}{}                              & Teacher (ResNet-18)  & 71.0          & 81.2          & 88.7               & 67.8          & 79.1          & 87.9           \\ \hline 
\end{tabular}}
\end{table}

\subsection{Cross Quality Distillation} 

As an additional experiment we do the distillation of embeddings to transfer knowledge between different domains. This was originally proposed in a classification setting by Su et al.~\cite{su2016adapting} who, in order to improve the recognition on low-quality data, use distillation with a teacher trained with high-quality data. The student then is trained with the low-quality data and the guidance from the teacher which has access to the high-quality data. Here we will apply cross quality distillation with the proposed losses for metric learning.
Since in this experiment the objective is not to reduce the number of parameters but to bridge a domain gap, we use the same architecture (ResNet-18) for the teacher and the student. To train the embeddings we use triplet loss and, as in the previous experiments, we train the students with two teachers: relative and absolute. 

We consider two cross quality distillation experiments on CUB-200-2011 and Cars-196. The first experiment considers \textit{low and high-resolution} images. To get the low-resolution images, we downsample them to 50 x 50 and then upsample them again to 224 x 224 (see examples in Fig.~\ref{fig:dataset}). The second experiment considers \textit{unlocalized signal degradation}, where the input images are cropped according to the given bounding boxes for the teachers, but not cropped for the students. The results can be seen in Table~\ref{tab:cqd}.

It can be seen that incorporating the additional knowledge distillation losses improves the results for most settings, with the relative teachers consistently surpassing the absolute ones, as in the previous experiments. The improvement by the distillation is more noticeable on the Cars-196 dataset which is also observed in~\cite{su2016adapting}. Since it is a more challenging dataset which has cars with different colors belonging to the same category, the information provided by the teacher becomes more crucial.

\section{Conclusions}\label{sec:conc}
We have investigated network distillation with the aim of computing efficient image embedding networks. We have proposed two losses with the aim to communicate the teacher network knowledge to the student network. We evaluate our approach on several datasets, and report significant improvements: we obtain a 6.3\% gain on CUB-200-2011, a 29.9\% gain on Cars-196 and a 6.3\% gain on Stanford Online Products for Recall@$1$ when compared to a student network of the exact same capacity which was trained without a teacher network. Furthermore, we apply our distillation loss to MobileNet-0.25. It greatly improves the Recall@1 by 17.1\%. We also verify the benefit of adding unlabeled data for embedding learning. In addition, we demonstrate that an embedding learned on high-quality images can be used to improve the student network which has only access to low quality images.

\minisection{Acknowledgement}
This work was supported
by TIN2016-79717-R of the Spanish Ministry, the CERCA
Program and the Industrial Doctorate Grant 2016 DI 039 of the Ministry of Economy and Knowledge of the Generalitat de Catalunya. Xialei Liu acknowledges the Chinese Scholarship Council (CSC) grant No.201506290018. We also acknowledge the generous GPU support from NVIDIA.

{\small
\bibliographystyle{ieee}
\bibliography{longstrings,egbib}
}

\end{document}